\documentclass[
]{ceurart}

\usepackage{subcaption}
\begin{document}

%%
%% Rights management information.
%% CC-BY is default license.
\copyrightyear{2023}
\copyrightclause{Copyright for this paper by its authors.
  Use permitted under Creative Commons License Attribution 4.0
  International (CC BY 4.0).}

%%
%% This command is for the conference information
\conference{HITLAML’23: Human-in-the-Loop Applied Machine Learning, September 04–06, 2023, Belval, Luxembourg}

%%
%% The "title" command
\title{Unveiling the Role of Expert Guidance: A Comparative Analysis of User-centered Imitation Learning and Traditional Reinforcement Learning}

%%
%% The "author" command and its associated commands are used to define
%% the authors and their affiliations.
\author[1,2]{Amr Gomaa}[%
orcid=0000-0003-0955-3181,
email=amr.gomaa@dfki.de,
url=https://amrgomaaelhady.github.io/,
]

\author[1]{Bilal Mahdy}[%
email=bilal.mahdy@dfki.de,
]

\address[1]{German Research Center for Artificial Intelligence (DFKI), Saarbr{\"u}cken, Germany}
\address[2]{Saarland Informatics Campus, Saarland University, Saarbr{\"u}cken, Germany}

%%
%% The abstract is a short summary of the work to be presented in the
%% article.
\begin{abstract}
Integration of human feedback plays a key role in improving the learning capabilities of intelligent systems. This comparative study delves into the performance, robustness, and limitations of imitation learning compared to traditional reinforcement learning methods within these systems. Recognizing the value of human-in-the-loop feedback, we investigate the influence of expert guidance and suboptimal demonstrations on the learning process. Through extensive experimentation and evaluations conducted in a pre-existing simulation environment using the Unity platform, we meticulously analyze the effectiveness and limitations of these learning approaches. The insights gained from this study contribute to the advancement of human-centered artificial intelligence by highlighting the benefits and challenges associated with the incorporation of human feedback into the learning process. Ultimately, this research promotes the development of models that can effectively address complex real-world problems.
\end{abstract}

%%
%% Keywords. The author(s) should pick words that accurately describe
%% the work being presented. Separate the keywords with commas.

%%
%% This command processes the author and affiliation and title
%% information and builds the first part of the formatted document.
\maketitle

\section{Introduction and Related Work}

Human-centered artificial intelligence (HCAI) is an exciting new area of research that is attracting increasing attention from researchers of both artificial intelligence (AI) and human-computer interaction (HCI)~\cite{xu2019toward,nowak2018assessing,bryson2019society,shneiderman2020human}. Despite the significant progress made in developing autonomous systems, these systems still rely heavily on human operators, local or remote, to intervene and help or take control in situations where the system cannot proceed, highlighting the need for HCAI techniques to promote trust, control, and reliability between users and machines~\cite{shneiderman2020human}. However, developing and implementing these concepts remains a challenging and complex task~\cite{nowak2018assessing}. As a result, there is still much room for improvement and further research in this field~\cite{bryson2019society}. Several approaches have proposed ways to incorporate human knowledge into neural networks as a way of initialization, to guide network refinement, and to extract symbolic information from the network~\cite{shavlik_combining_1994, von2019informed}. More recent attempts have tried to combine deep learning with knowledge bases in joint models (e.g., for construction and population)~\cite{ratnerAlexEtAl2018, adel2018deep}. Some work has focused on integrating neural networks with classical planning by mapping subsymbolic input to symbolic one, which automatic planners can use~\cite{asai2018classical}. 

Recently, reinforcement learning (RL) \cite{Sutton1998} has reemerged as a promising machine learning approach within the field of autonomous systems (e.g., ChatGPT). These methods have demonstrated increasing effectiveness in optimizing reward functions for complex environments. However, shaping appropriate reward functions for intricate tasks and encompassing their aspects remains a challenge \cite{hadfieldmenell2020inverse}. In contrast, humans excel at rapidly acquiring complex skills by observing and imitating others. Similarly, autonomous agents can take advantage of this concept, known as learning from demonstrations (LfD) \cite{argal_2009}, to address the challenges mentioned above using imitation learning (IL) methods using expert demonstrations \cite{finn2016guided}. Behavioral cloning (BC) \cite{ros2011} and Generative Adversarial Imitation Learning (GAIL) are the state-of-the-art and most prominent approaches employed to tackle imitation learning problems where the agent has access to state and action information from the demonstrations \cite{ho2016generative}. 
Significant progress has been made in Reinforcement Learning (RL) and Imitation Learning (IL) domains. Torabi et al. \cite{torabi2018behavioral} introduced an advanced adaptation of behavioral cloning known as Behavioral Cloning from Observation, where the agent solely observes demonstration states without access to the corresponding demonstration actions. In a separate study by Taylor \cite{ijcai2018-817}, several methods were proposed to facilitate the agent's optimal utilization of knowledge from suboptimal human demonstrations, including Learning from Human Demonstrations and Learning from Human Feedback. Fang et al.~\cite{fang2022target} compared reinforcement and imitation learning for indoor visual navigation. Unlike previous works, ours focuses solely on analyzing the efficacy of imitation learning techniques to assess the importance of learning from demonstrations as a human-in-the-loop learning paradigm in a highly complex environment, regardless of the application domain. 

Thus, \textbf{this paper contributes to the field of imitation and reinforcement learning, evaluating its performance, robustness, and limitations}. We conduct a detailed investigation into the performance of these state-of-the-art imitation learning techniques in the context of a simulated Bird Hunter game using \textit{Unity ml-agents}\footnote{\url{https://github.com/Unity-Technologies/ml-agents}} and \textit{Pytorch}\footnote{\url{https://pytorch.org/}} to evaluate and compare their effectiveness with traditional RL techniques; we investigate the impact of expert guidance and suboptimal demonstrations on imitation learning performance compared to traditional reinforcement learning in diverse environmental complexities. We utilize the Proximal Policy Approximation (PPO)~\cite{schulman2017proximal} and the Soft-Actor Critic (SAC)~\cite{haarnoja2018soft} methods for our investigation as the most used reinforcement learning techniques, especially in simulation frameworks such as Unity. We provide valuable insights into the comparative efficacy of IL and traditional RL, contributing to the development of intelligent systems in various environmental contexts.

\section{Methodology}

Our study adopts a systematic and progressive approach to comprehensively evaluate the effectiveness of imitation learning with suboptimal and expert demonstrations, as well as its comparison to reinforcement learning techniques such as PPO and SAC. Incremental complexities are introduced to the base environment, incorporating new parameters and analytical challenges at each stage, such as transitioning from grayscale to a colored environment and introducing various bird species with distinct reward schemes. In reinforcement learning, the agent interacts with an environment by selecting actions and receiving feedback through observations and rewards. The observations provide information about the current state of the environment, while the rewards serve as feedback signals that indicate the desirability of the agent's actions. Therefore, for each level of environment complexity, we establish the states of observation and action, as well as the corresponding reward structure (i.e., reward shaping).

\begin{figure}[t]
    % \hspace{-5cm}
    \begin{subfigure}{0.3\linewidth}
            \centering
            \hspace{-2cm}
            \includegraphics[width=0.55\linewidth,keepaspectratio]{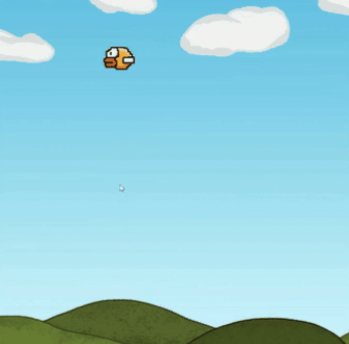}
        \end{subfigure}
    \begin{subfigure}{0.3\linewidth}
            \centering
            \hspace{-2.9cm}
            \includegraphics[width=0.5\linewidth,keepaspectratio]{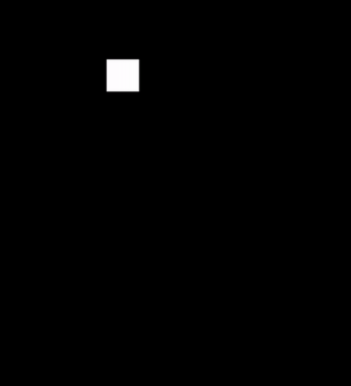}
        \end{subfigure}
    \begin{subfigure}{0.3\linewidth}
            \centering
            \hspace{-2cm}
            \includegraphics[width=\linewidth,keepaspectratio]{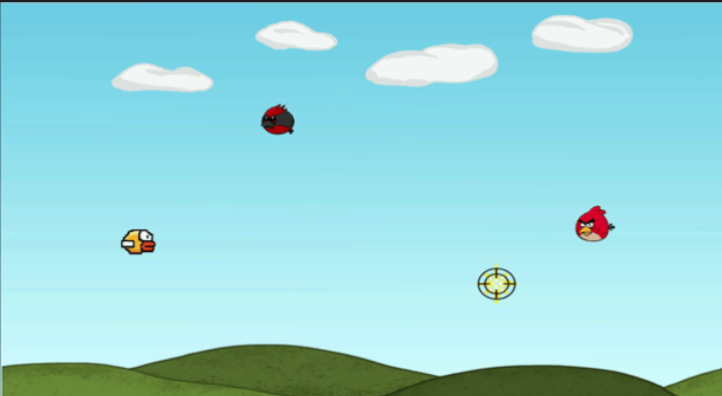}
        \end{subfigure}
        \caption{Screenshots (not to scale) of the Bird Hunter game showing the base environment (left), the grayscale backdrop camera view (middle), and the high-complexity environment (right).}
        \label{fig:img_multi_bird}
\end{figure}

\paragraph{Base Environment (Low-complexity Environment).} We conducted our study using a preexisting 2D simulated Bird Hunter game to train an autonomous agent (see~\autoref{fig:img_multi_bird}). Initially, a grayscale backdrop was used, with the bird represented as a white box on a black background. The camera sensor captured grayscale images at a resolution of 50 pixels for each axis (x and y), resulting in an observation space of 2500 pixels (50 x 50 x 1), where the one represents a single channel image. The agent's actions involved discrete pixel coordinate pairs for movement, with shooting performed automatically and not treated as a separate action. The reward function (as seen in~\autoref{eq:grayreward}) assigns a reward of (+1) for hitting a bird and a negative reward of (-0.01) for missing.

        \begin{equation}\label{eq:grayreward}
        RewardFunction = \left\{
                    \begin{array}{ll}
                        +1 & \quad Bird\:hit \\
                        -0.01 & \quad Bird\:missed
                    \end{array}
                \right.
        \end{equation}

\paragraph{Colored Environment (Medium-complexity Environment).} This setting builds on the initial environment by utilizing RGB color channels instead of grayscale for the background. The scaling and action space remain the same as in the base environment, while the observation space expands to 7500 pixels (50 x 50 x 3), with the three representing the color channels. The reward function remains consistent with that of the base environment.

\paragraph{Limited Ammunition with Multiple Bird Environment (High Complexity Environment).} In this environment, we enhance the complexity by assigning meaning to the colors in the agent's observation, rather than simply introducing a color channel to the environment. In addition to the existing yellow bird as primary target, two new types of birds are introduced. The red bird serves as a bonus, appearing when the agent successfully hits two yellow birds, while the black bird acts as a bomb, exploding upon contact (see~\autoref{fig:img_multi_bird}). Consequently, the reward function is updated to include additional rewards for the red bird (+2) and the black bird (-0.5) as seen in~\autoref{eq:rf_limited_ammo}. 

        \begin{equation}\label{eq:rf_limited_ammo}
        RewardFunction = \left\{
                    \begin{array}{ll}
                        +1 & \quad Yellow\:Bird\:hit \\
                        +2 & \quad Red\:Bird\:hit \\
                        -0.01 & \quad Bird\:missed \\
                        -0.5 & \quad Black\:Bird\:hit
                    \end{array}
                \right.
        \end{equation}

Furthermore, we introduce new parameters to enhance the agent's convergence towards pinpoint accuracy. The parameter $Clip_{Size}$ is introduced to determine a preset amount of ammunition available for shooting. Another virtual-dependent parameter, $Ammo_t$, specifies the ammunition available to the player at time $t$. Furthermore, the duration of reload $T_{reload}$ is incorporated to determine the time steps required to complete a reload action.
At each time step $t$, if ammunition is available ($Ammo_t > 0$), the agent is compelled to shoot. Otherwise, a reload action is enforced, resetting the ammunition available to $Clip_{Size}$ after $T_{reload}$ time steps as seen in~\autoref{eq:ammo}.

        \begin{equation}\label{eq:ammo}
        Ammo_t = \left\{
                    \begin{array}{ll}
                        Ammo_{t-1}-1 & \quad Ammo_{t-1} > 0 \\
                        Clip\_Size & \quad t\:(\mathrm{mod}{(Clip\_Size + T_{reload})}) = 0 \\
                        0 & \quad otherwise
                    \end{array}
                \right.
        \end{equation}

\section{Discussion and Results}

In this section, we present the results obtained from different environment settings using various RL and IL approaches. The comparison between approaches in each respective environment is based on the evaluation metrics traditionally used to assess RL agents, as outlined below:

\begin{itemize}
    \item \textit{Cumulative Reward Function}: The mean reward obtained by the agent in a specified number of steps. Higher value indicates better performance.
    \item \textit{Episode length}: The time taken for the agent to complete an episode, where episodes end when any bird is shot. Lower value indicates better performance.
    \item \textit{Entropy}: A measure of the agent's uncertainty in choosing an action given the observed state. Lower value indicates better performance.
\end{itemize}

\begin{figure}[t]
    % \hspace{-5cm}
    \begin{subfigure}{0.4\linewidth}
            \centering
            \includegraphics[width=0.96\linewidth,keepaspectratio]{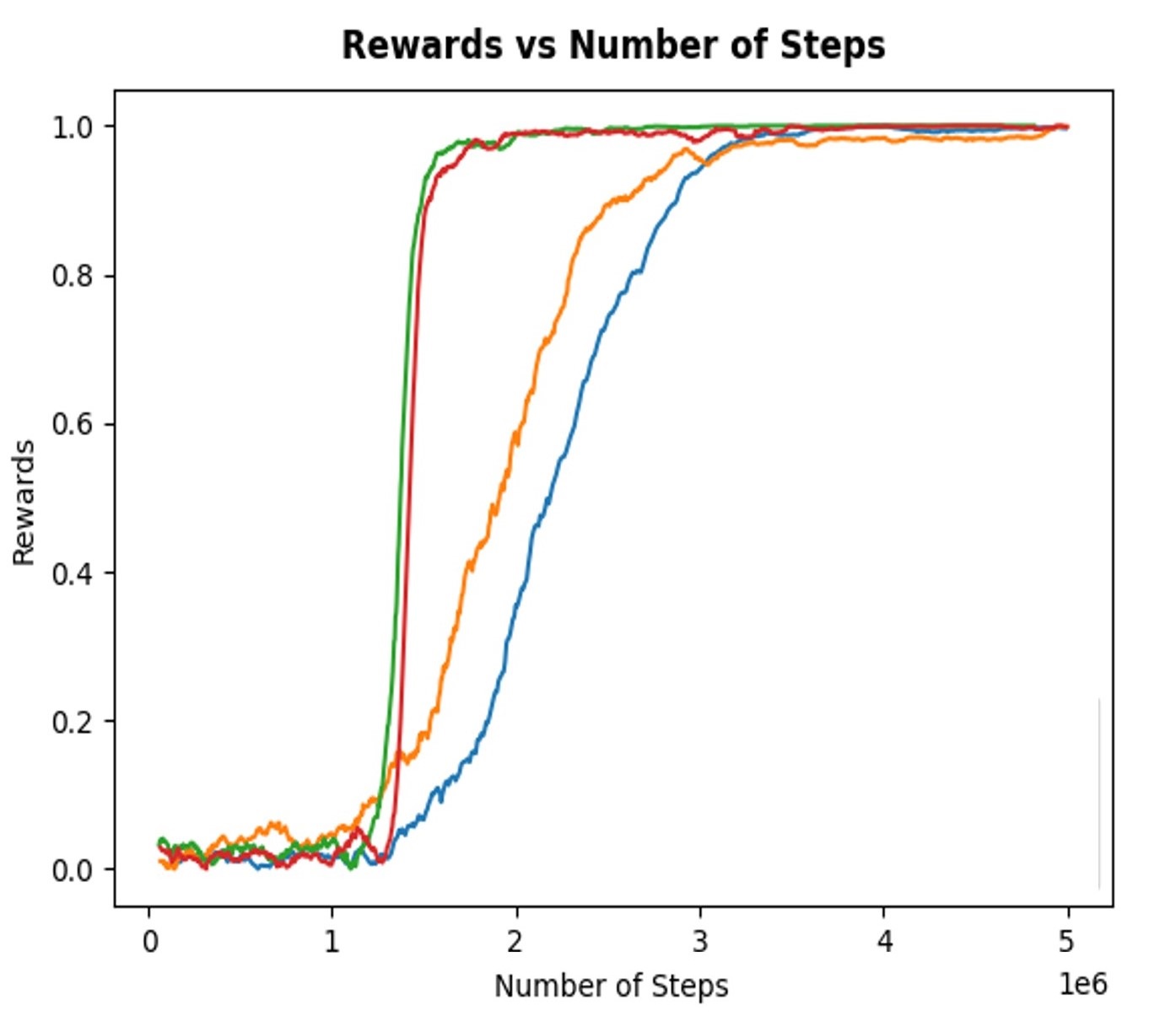}
        \end{subfigure}
    \begin{subfigure}{0.4\linewidth}
            \centering
            \includegraphics[width=\linewidth,keepaspectratio]{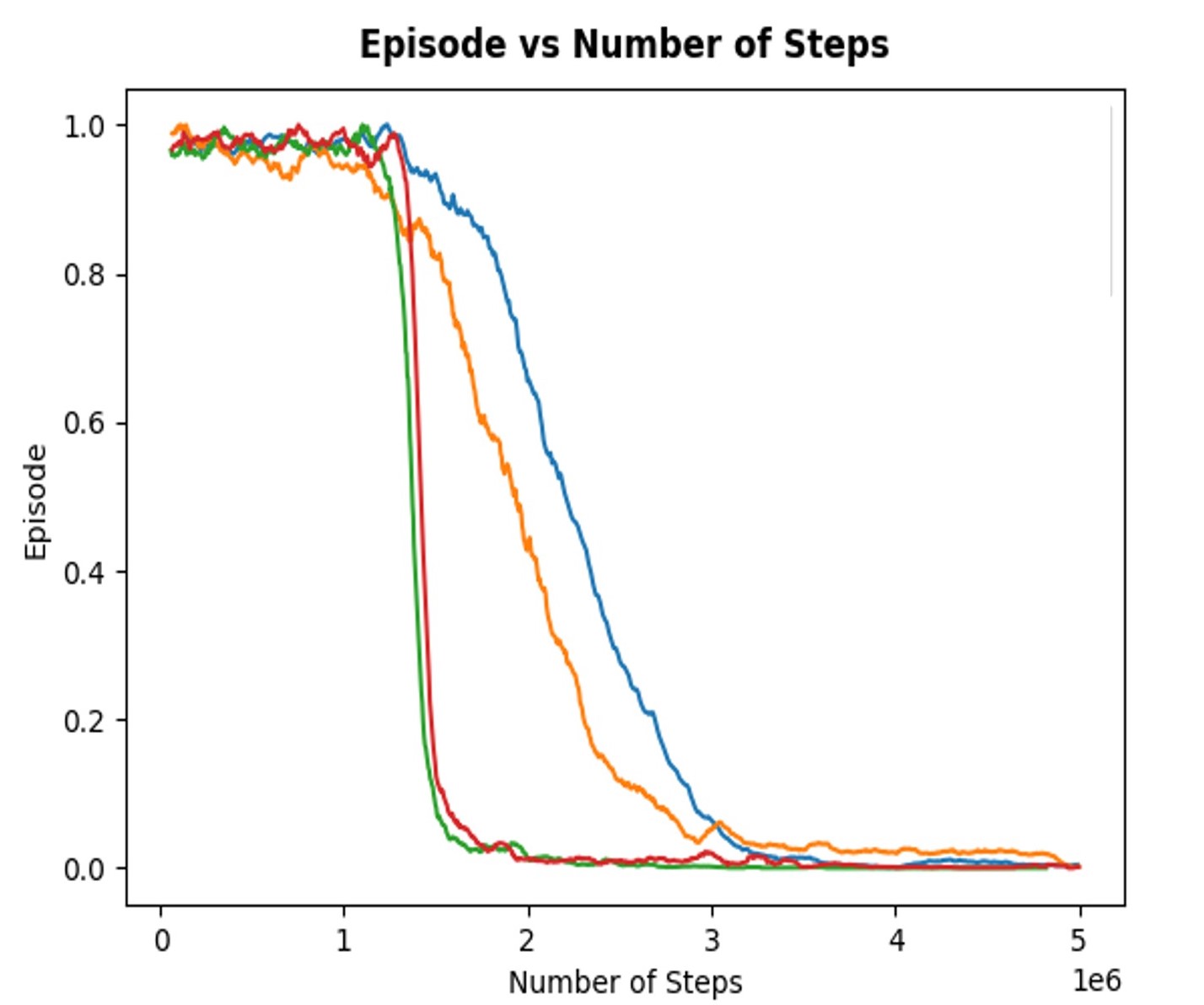}
        \end{subfigure}
    \centering \begin{subfigure}{0.4\linewidth}
        \vspace{0.2cm}
            \centering
            \includegraphics[width=0.96\linewidth,keepaspectratio]{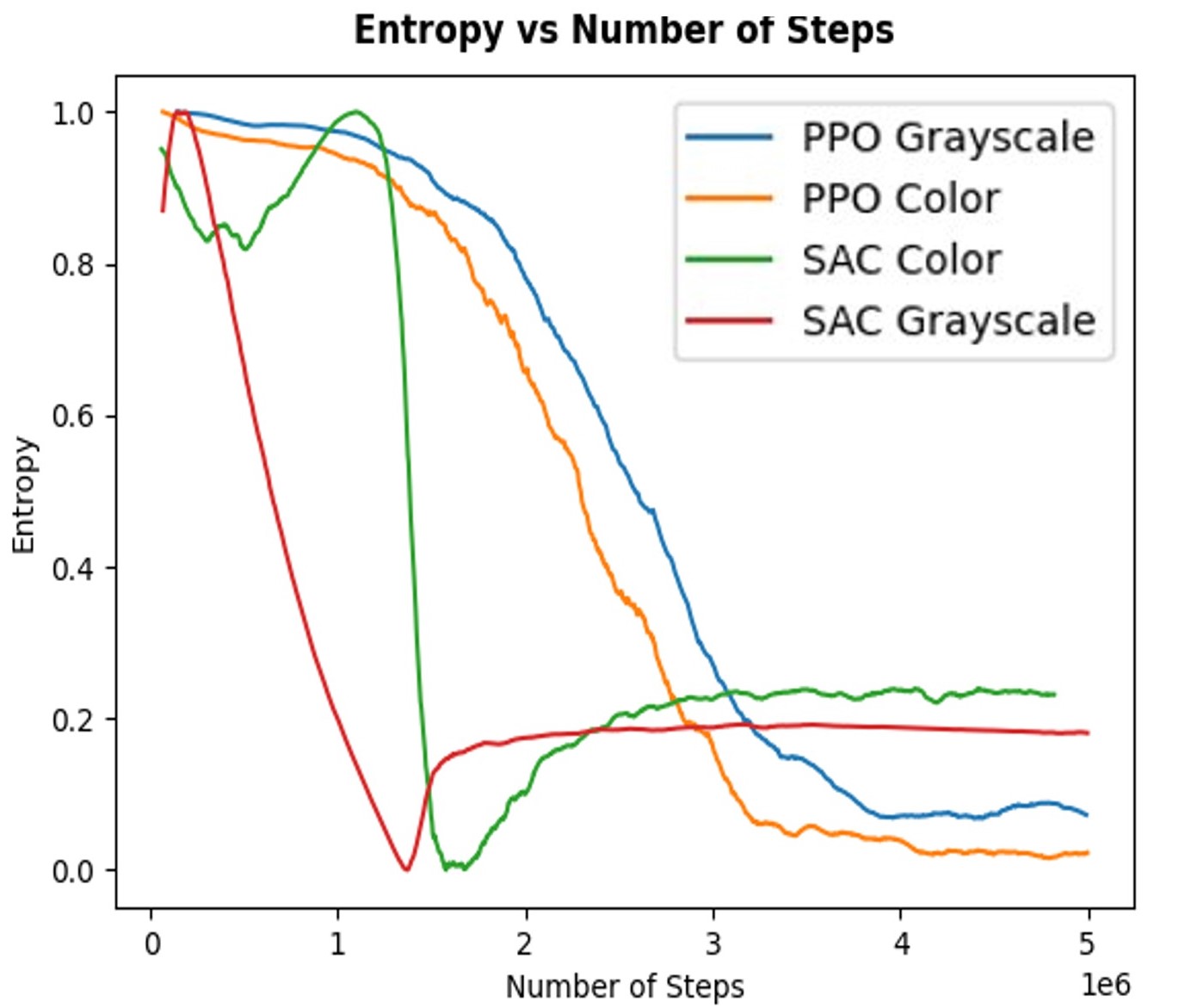}
        \end{subfigure}
        \caption{Comparing base environment with the RGB environment for different RL algorithms.}
        \label{fig:ppo_sac_first}
\end{figure}

\begin{figure}[b]
    % \hspace{-5cm}
    \begin{subfigure}{0.4\linewidth}
            \centering
            \includegraphics[width=\linewidth,keepaspectratio]{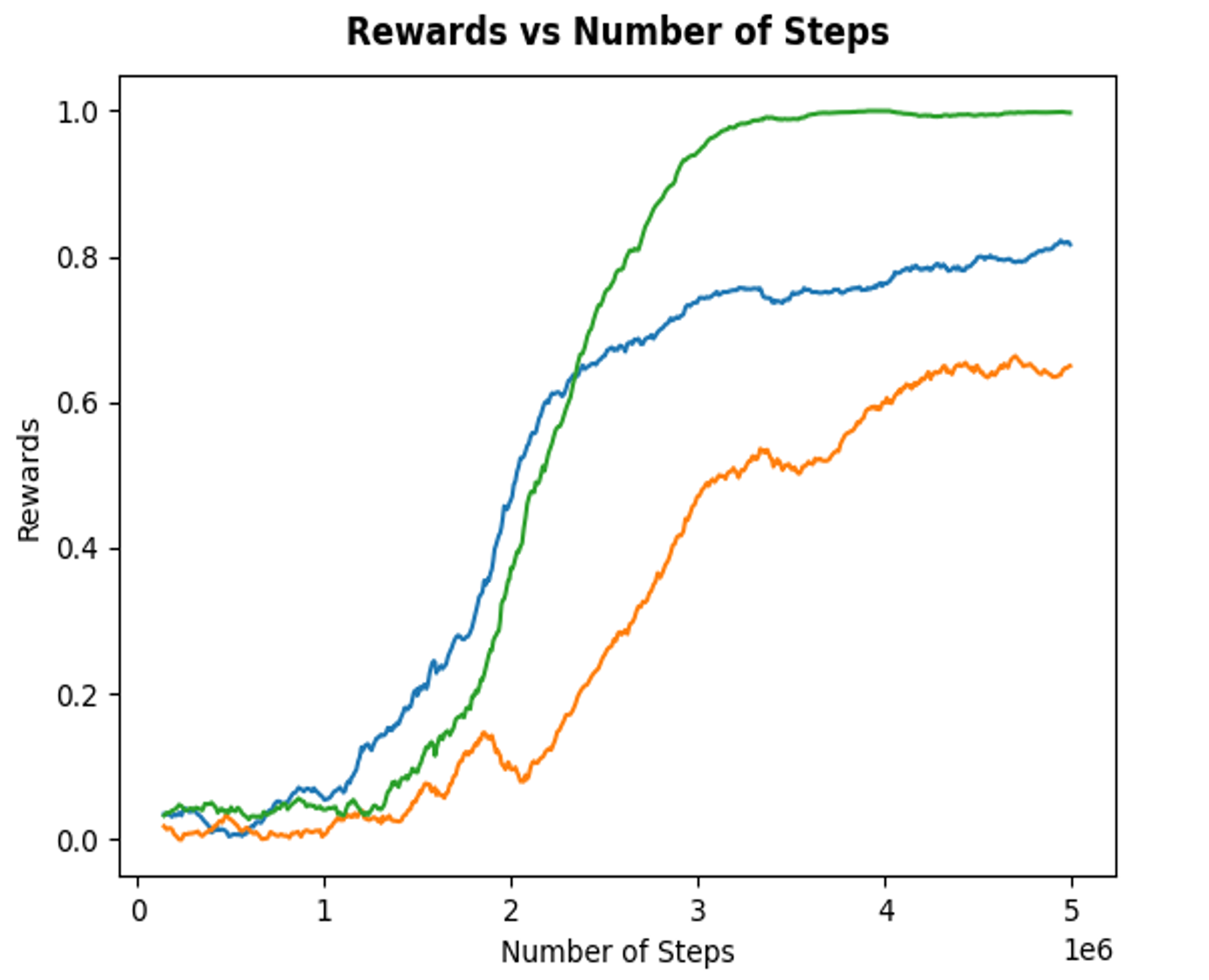}
        \end{subfigure}
    \begin{subfigure}{0.4\linewidth}
            \centering
            \includegraphics[width=\linewidth,keepaspectratio]{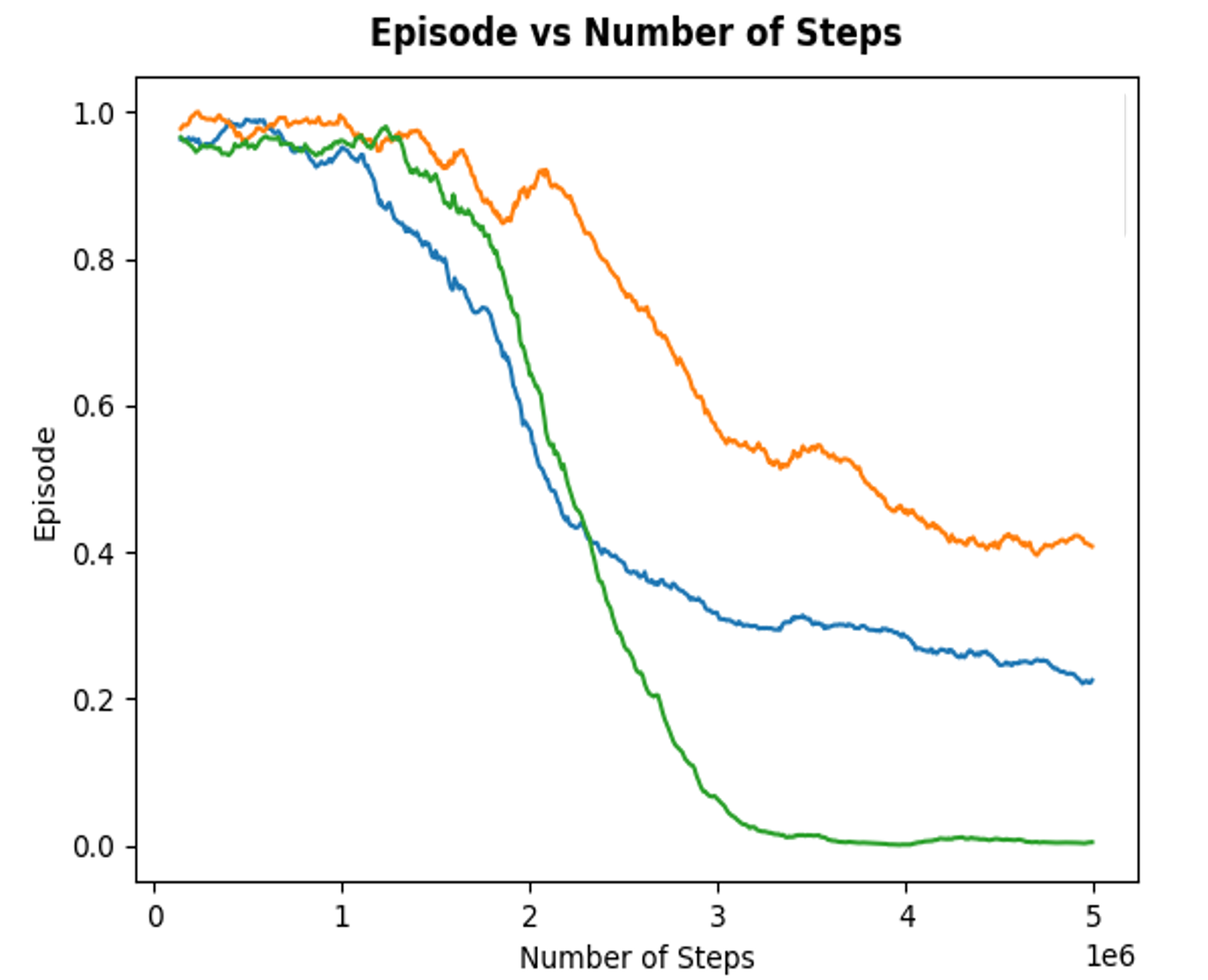}
        \end{subfigure}
    \centering \begin{subfigure}{0.4\linewidth}
            \centering
            \vspace{0.2cm}
            \includegraphics[width=\linewidth,keepaspectratio]{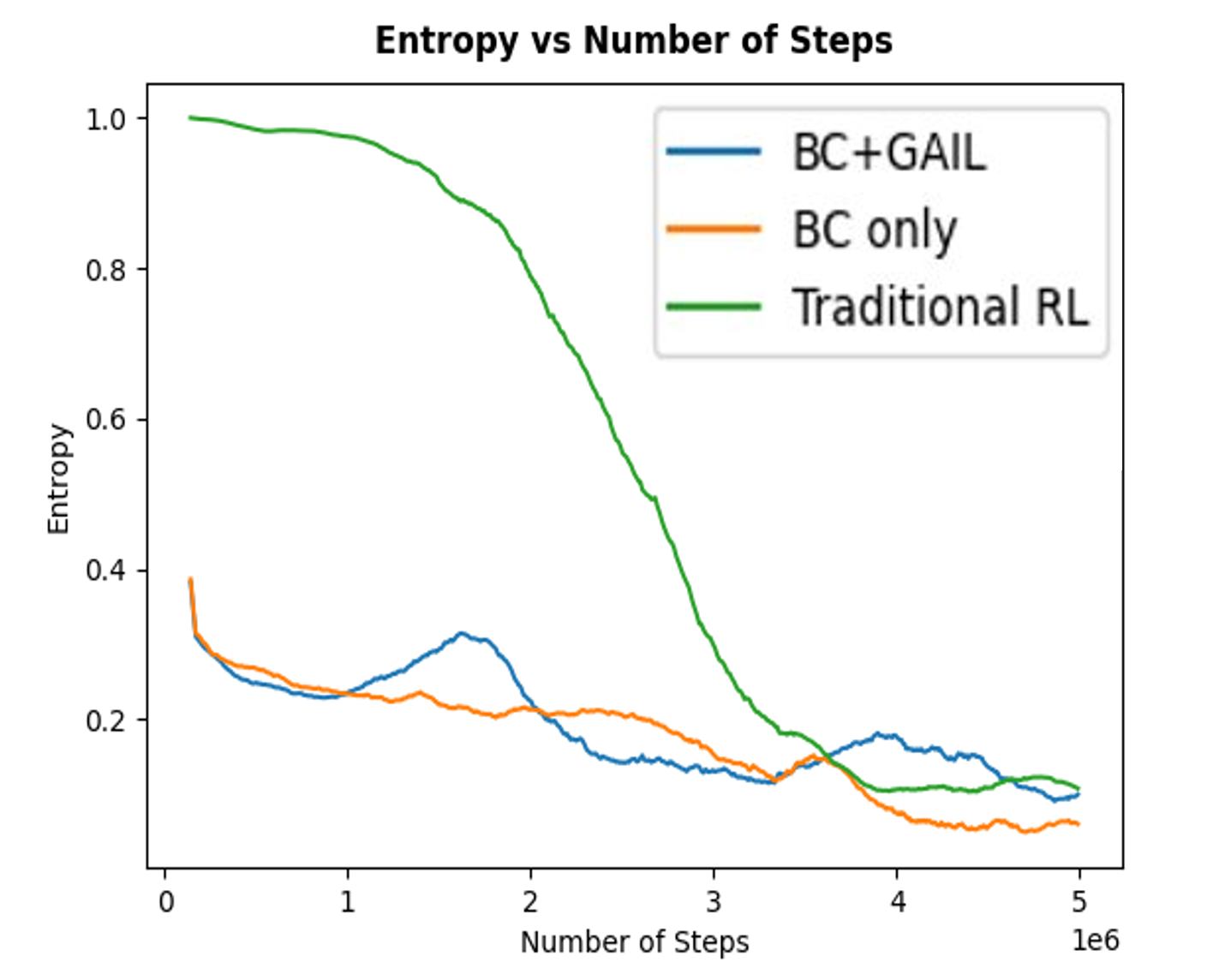}
        \end{subfigure}
        \caption{Comparing different IL algorithms in the RGB environment.}
        \label{fig:rl_il_first}
\end{figure}

\subsection{Low and Medium Complexity Setting}

First, we compare the performance of both SAC and PPO RL algorithms for the grayscale environment and the RGB environment (i.e., low vs medium complex environments), then choose the superior RL algorithm to compare RL to IL approaches.~\autoref{fig:ppo_sac_first} shows the comparison of the RL algorithms in terms of the metrics mentioned above. While PPO's entropy is lower than that of SAC, indicating a relatively more stable choice of actions, SAC converged faster than PPO in terms of cumulative reward and step count. Thus, SAC is used in further comparisons. Next, we compare traditional RL algorithms (i.e., SAC) to IL techniques (i.e., BC and GAIL).~\autoref{fig:rl_il_first} and~\autoref{tab:RL-IL-table} show the results for the RL and IL comparison. It can be seen that while RL converges faster than both BC and GAIL, the latter IL techniques have a better entropy, indicating more stable learning and consistent action choices. It is also noticed that using the GAIL technique alone is not stable and hard to converge for this medium complexity environment, even for training for a very long number of steps (i.e., greater than million steps).

\begin{figure}[b]
    % \hspace{-5cm}
    \begin{subfigure}{0.4\linewidth}
            \centering
            \includegraphics[width=\linewidth,keepaspectratio]{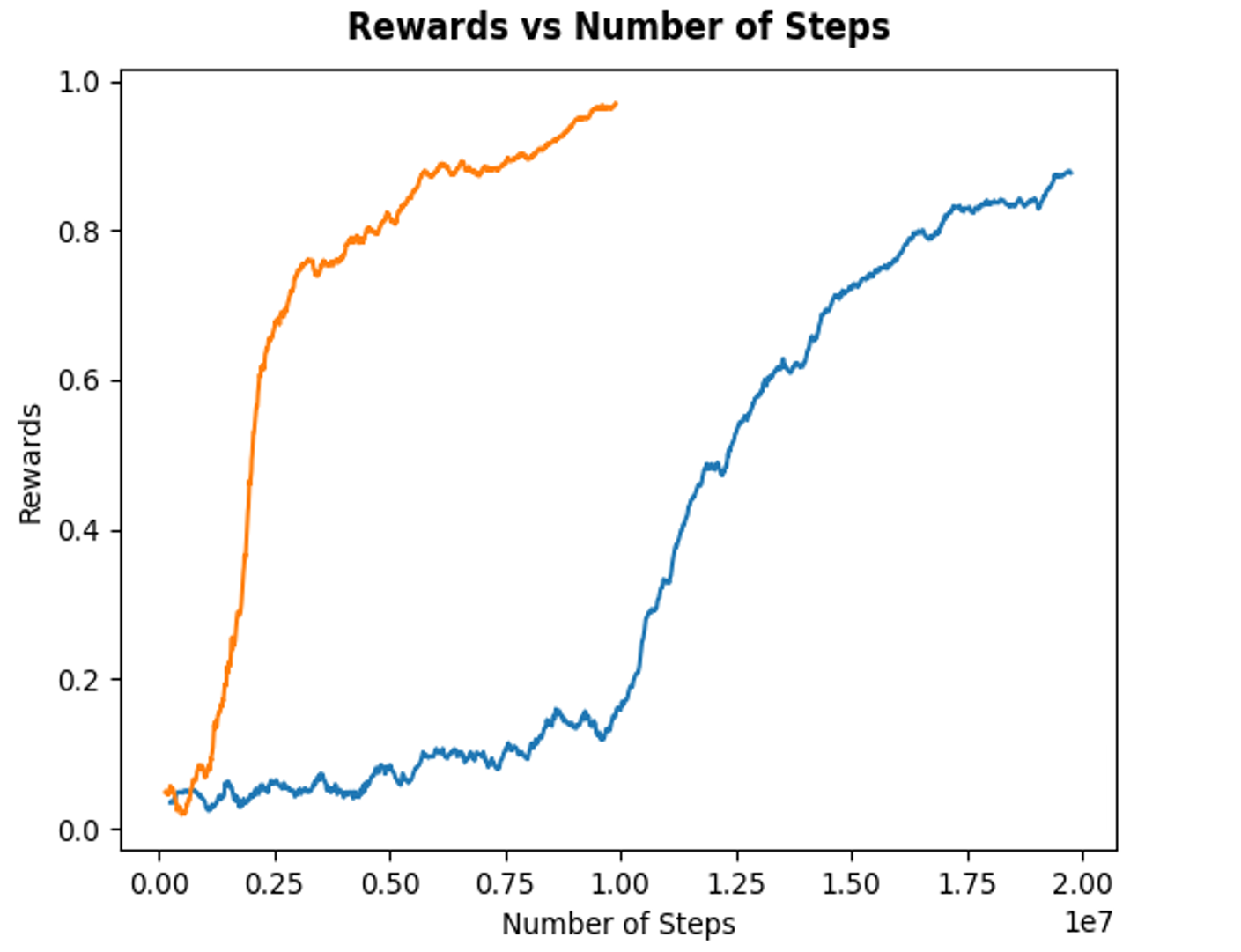}
        \end{subfigure}
    \begin{subfigure}{0.4\linewidth}
            \centering
            \includegraphics[width=\linewidth,keepaspectratio]{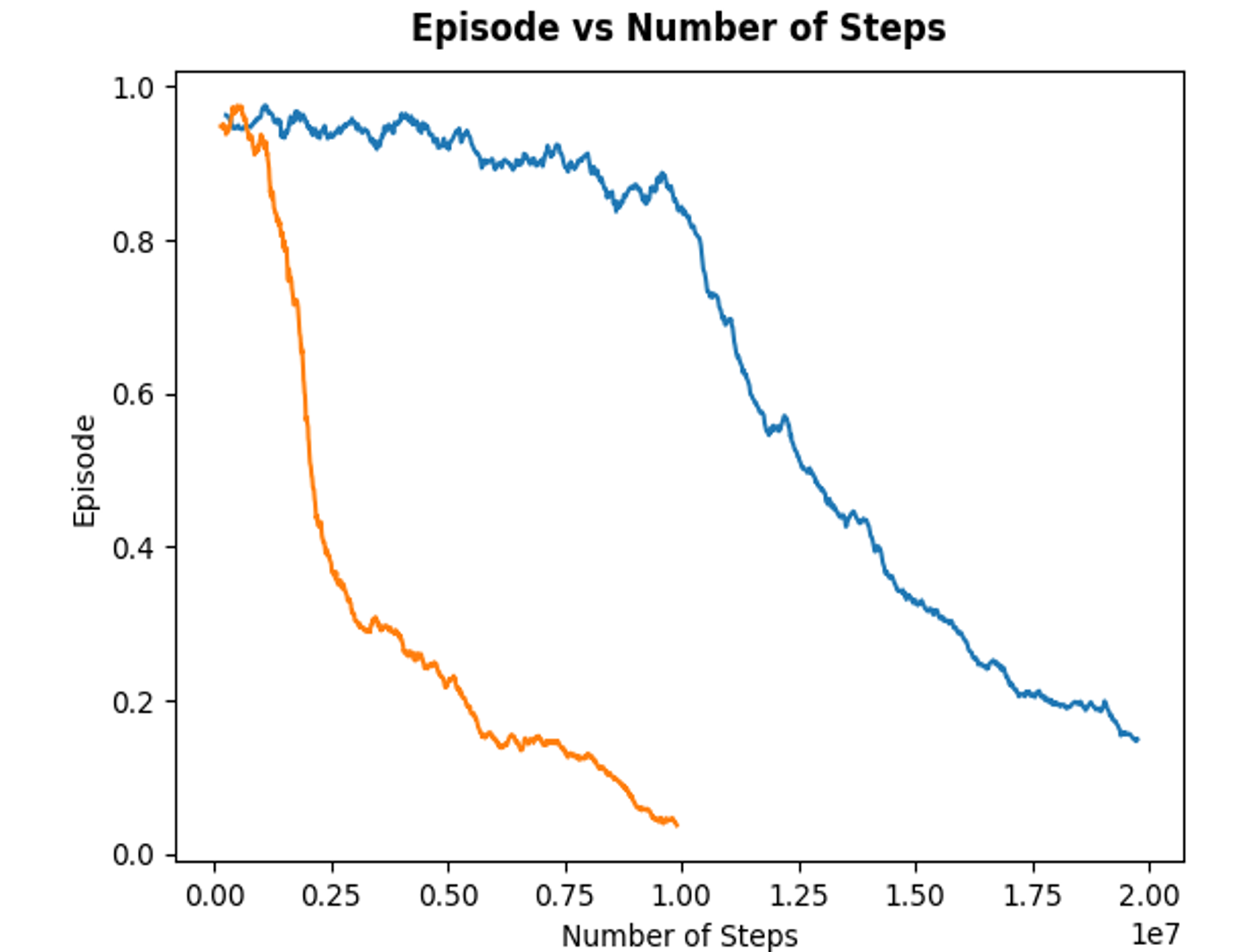}
        \end{subfigure}
    \centering \begin{subfigure}{0.4\linewidth}
            \centering
            \vspace{0.2cm}
            \includegraphics[width=\linewidth,keepaspectratio]{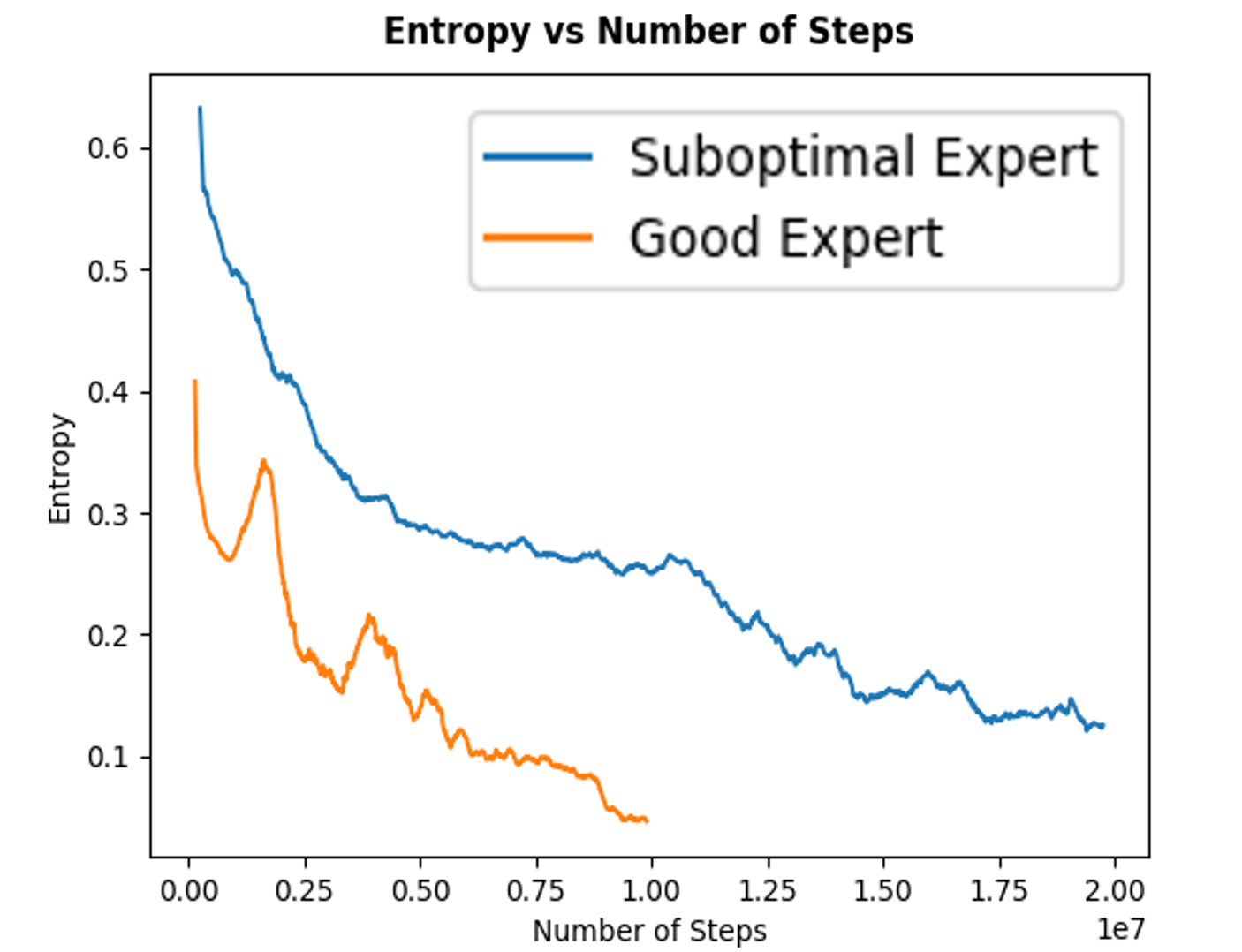}
        \end{subfigure}
        \caption{Comparing imitation learning (BC + GAIL) for a good demonstration (i.e., expert user) and suboptimal demonstration (i.e., novice user).}
        \label{fig:expert_demo}
\end{figure}

Lastly, the RGB environment was evaluated by comparing two types of demonstrations used to train imitation learning techniques (BC + GAIL): one from a proficient experienced user and the other from a suboptimal user. Both demonstrations came from the same user to ensure consistency, where the user attempted the shoting as best as he could for the expert demonstration and intentially missed few birds to record the demonstration of the suboptimal user. As a manipulation check, examination of the reward function showed that the competent expert performed the task with high accuracy, achieving a mean reward of $0.997$ with no missed shots. On the other hand, the suboptimal demonstration had a mean reward of $0.81$, indicating a higher frequency of missed shots.
These demonstrations aimed to evaluate the performance of imitation learning under the same environment complexity and conditions.~\autoref{fig:expert_demo} illustrates that the agent trained with the expert demonstration exhibited faster learning, greater consistency, and more confident action selection. In contrast, the agent trained with the suboptimal demonstration eventually converged, but it took twice as long as the expert demonstration.

\begin{table*}
  \caption{Comparison of Traditional RL and IL methods in the ``Base'' environment. The maximum reward achievable by the agent is one.}
    \label{tab:RL-IL-table}
  \begin{tabular}{c|cccc}
    \toprule
    \textbf{Metric} & RL (SAC) & BC Only & GAIL Only & BC \& GAIL \\
    \midrule
    \textbf{Step Count} & \textbf{162k} & $>$500k & $\gg$1M & $\sim$500k \\
         % \hline
    \textbf{Cumulative Reward} & \textbf{0.98} & 0.89 & No Convergence & 0.95 \\
         % \hline
    \textbf{Entropy} & 0.66 & \textbf{0.45} & No Convergence & 0.63 \\
  \bottomrule
\end{tabular}
\end{table*}

\begin{figure}[t]
    % \hspace{-5cm}
    \begin{subfigure}{0.4\linewidth}
            \centering
            \includegraphics[width=\linewidth,keepaspectratio]{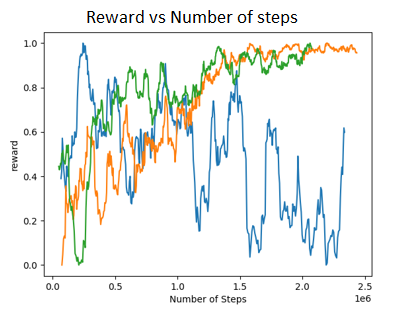}
        \end{subfigure}
    \begin{subfigure}{0.4\linewidth}
            \centering
            \includegraphics[width=\linewidth,keepaspectratio]{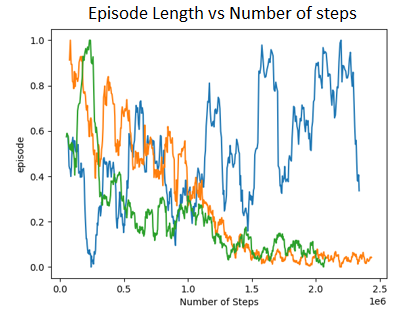}
        \end{subfigure}
    \centering \begin{subfigure}{0.4\linewidth}
            \centering
            \vspace{0.2cm}
            \includegraphics[width=\linewidth,keepaspectratio]{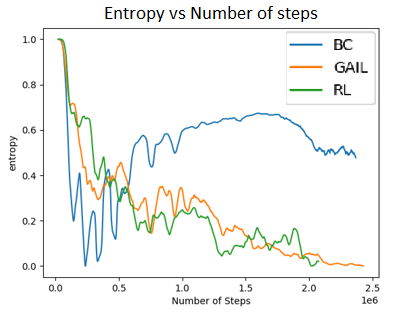}
        \end{subfigure}
        \caption{Comparison of average cumulative reward, episode length and model entropy between RL, BC and GAIL in the Limited Ammo environment}
        \label{fig:limited_ammo_results_fig}
\end{figure}

\subsection{High Complexity Setting}

In order to further assess the performance of the GAIL, BC, and RL algorithms, we performed evaluations in a highly complex environment. This environment included multiple birds with different rewards and limited ammunition, as described in the Methods section. Building on the insights gained from the previously mentioned evaluations of imitation learning techniques, we modified the training approach for BC and GAIL. Instead of relying solely on demonstrations, these algorithms were trained with a combination of intrinsic and extrinsic rewards. This adjustment was made to address the tendency of BC and GAIL to deviate from an optimal policy when trained with demonstrations only. The RL and IL comparison results in this highly complex environment are presented in~\autoref{fig:limited_ammo_results_fig} and~\autoref{tab:limited_ammo_results}, which provide a comparison of the RL, BC, and GAIL algorithms. These results offer insight into the performance and effectiveness of each algorithm in this challenging setting.

\begin{table*}
  \caption{Comparison between RL, BC and GAIL in the ``Limited Ammo and Multiple Birds'' environment. The maximum reward that the agent can achieve is two. The expert demo had an average reward of \textbf{1.5}.}
    \label{tab:limited_ammo_results}
  \begin{tabular}{c|ccc}
    \toprule
    \textbf{Metric} & RL & BC & GAIL \\
    \midrule
    \textbf{Convergence Step Count} & 2.5M & No Convergence & \textbf{1M} \\
         % \hline
    \textbf{Cumulative Reward} & 1.4 & -0.77 & \textbf{1.67} \\
         % \hline
    \textbf{Entropy} & 0.68 & 6.64 & \textbf{0.23} \\
  \bottomrule
\end{tabular}
\end{table*}

\textit{Traditional RL.}
In this highly complex environment, the traditional RL algorithm failed to capture an effective bird-shooting strategy. Instead, it resorted to ``cheating'' the environment by learning the average spawn locations of the red and yellow birds. The agent then focused solely on shooting at these specific spots, barely moving the cursor. Remarkably, the traditional RL agent achieved a score close to that of a human player using this method. This highlights the ability of RL algorithms to exploit loopholes given sufficient time.

\textit{Behavioural Cloning.}
In contrast, the BC algorithm encountered significant difficulties in achieving the score of a human player. Since the recorded demonstration did not utilize the environment loophole but instead moved the cursor around and aimed at the red and yellow birds while avoiding the black ones, the agent struggled to replicate the demonstrated behavior and failed to converge or show improvement after a substantial number of iterations. This underscores the limitations of imitation learning algorithms relying solely on expert demonstrations and their reduced capacity for exploratory behavior compared to traditional RL.

\textit{GAIL.}
Initially, the GAIL algorithm faced similar challenges as the BC algorithm. However, due to its combined approach, GAIL was able to break free from recorded behavior and discover the same environment loophole exploited by the traditional RL algorithm. Ultimately, GAIL achieved the highest score among all algorithms, surpassing even the recorded human demonstrations, while achieving the lowest model entropy. This aligns with the notion that GAIL is particularly effective when dealing with environments of high complexity and dimensions.

\section{Conclusion and Future Work}

In conclusion, we compared policy optimization techniques and model architectures across various complexities of the environment, providing valuable information and avenues for future research. PPO demonstrated stable convergence and lower model entropy, indicating increased confidence in action selection. However, SAC exhibited superior sample efficiency and faster convergence, emphasizing the stability-efficiency trade-off, making it favorable when time is limited. The imitation learning algorithms converged slower but had a lower model entropy, relying heavily on expert demonstrations and limiting loophole exploitation. Traditional reinforcement learning algorithms discovered loopholes through reward-shaping complexity rather than learning intended behavior. GAIL performed well by effectively capturing expert demonstrations, achieving higher scores, and lower model entropy. This highlights the potential of imitation learning to overcome reinforcement learning limitations. On the other hand, reinforcement learning outperformed imitation learning in simple low-complexity environments where reward shaping is not challenging. Future research should explore performance in different domains, and develop hybrid approaches that take advantage of multiple algorithms to enhance convergence, stability, and exploration capabilities.

%%
%% The acknowledgments section is defined using the "acknowledgments" environment
%% (and NOT an unnumbered section). This ensures the proper
%% identification of the section in the article metadata, and the
%% consistent spelling of the heading.
\begin{acknowledgments}
This work is partially funded by the German Ministry of Education and Research (BMBF) under the TeachTAM project (Grant Number: 01IS17043) and the CAMELOT project (Grant Number: 01IW20008).
\end{acknowledgments}

%%
%% Define the bibliography file to be used
\bibliography{sample-ceur}

%%
%% If your work has an appendix, this is the place to put it.
% \appendix

\end{document}